\documentclass[conference]{IEEEtran}
\IEEEoverridecommandlockouts
\usepackage{cite}
\usepackage{amsmath,amssymb,amsfonts}
\usepackage{algorithmic}
\usepackage{graphicx}
\usepackage{textcomp}
\usepackage{xcolor}
\usepackage{hyperref}
\newcommand{\RN}[1]{\uppercase\expandafter{\romannumeral#1}}

\def\BibTeX{{\rm B\kern-.05em{\sc i\kern-.025em b}\kern-.08em
    T\kern-.1667em\lower.7ex\hbox{E}\kern-.125emX}}
\begin{document}

\title{Adversarial Defense in Vision-Language Models: An Overview\\

\thanks{This work was partially supported by National Natural Science Fund of China under Grants 92570110 and 62271090, Chongqing Natural Science Fund under Grant CSTB2024NSCQ-JQX0038, National Key R\&D Program of China under Grant 2021YFB3100800 and National Youth Talent Project.}
}

\author{\IEEEauthorblockN{Xiaowei Fu}
\IEEEauthorblockA{\textit{School of Microelectronics and Communication Engineering} \\
\textit{Chongqing University}\\
Chongqing, China \\
xwfu@cqu.edu.cn}
\and
\IEEEauthorblockN{Lei Zhang*}
\IEEEauthorblockA{\textit{School of Microelectronics and Communication Engineering} \\
\textit{Chongqing University}\\
Chongqing, China \\
leizhang@cqu.edu.cn*}
}

\maketitle

\begin{abstract}

The widespread use of Vision Language Models (VLMs, e.g. CLIP) has raised concerns about their vulnerability to sophisticated and imperceptible adversarial attacks. These attacks could compromise model performance and system security in cross-modal tasks. To address this challenge, three main defense paradigms have been proposed: Training-time Defense, Test-time Adaptation Defense, and Training-free Defense. Training-time Defense involves modifying the training process, typically through adversarial fine-tuning to improve the robustness to adversarial examples. While effective, this approach requires substantial computational resources and may not generalize across all adversarial attacks. Test-time Adaptation Defense focuses on adapting the model at inference time by updating its parameters to handle unlabeled adversarial examples, offering flexibility but often at the cost of increased complexity and computational overhead. Training-free Defense avoids modifying the model itself, instead focusing on altering the adversarial inputs or their feature embeddings, which enforces input perturbations to mitigate the impact of attacks without additional training. This survey reviews the latest advancements in adversarial defense strategies for VLMs, highlighting the strengths and limitations of such approaches and discussing ongoing challenges in enhancing the robustness of VLMs.
\end{abstract}


\section{Introduction}

In recent years, multi-modal models pre-trained on massive data have garnered widespread attention and applications, leading to significant developments within the AI community. Leveraging contrastive learning across an enormous quantity of image-text pairings, Visual Language Models (VLMs) have demonstrated superior performance in various zero-shot cross-modal tasks. Taking CLIP as an example, given a probe image and a set of text class labels, CLIP calculates the similarity between the visual embedding of the query image and the textual embeddings of candidate class labels, and predicts the class as the one with the highest similarity score. In this way, CLIP can significantly broaden the task boundaries of computer vision on unseen downstream tasks.

Despite this success, VLMs usually face severe challenges in terms of adversarial attacks. Recent studies have proved that adding tiny, imperceptible adversarial perturbations to the input image in image classification tasks can cause a sharp decline or even complete failure in the classification performance of these models. For example, a 1/255 perturbation budget is sufficient to cause predictions to collapse in CLIP. In practical applications, this vulnerability poses a serious challenge to safety-critical fields such as autonomous driving and medical diagnostics. Consequently, it has become an imperative research challenge to develop effective defense strategies for these advanced models.

To mitigate the adversarial vulnerability of VLMs, researchers have proposed various defense strategies~\cite{zhang2024meta,fu2025m3c}. Based on the processing stage and whether the model parameters change, these strategies can be summarized into three paradigms: Training-time defense, Test-time Adaptation defense, and Training-free defense. Training-time defense focuses on introducing adversarial examples during training and optimizing model parameters to enhance the robustness of the VLMs. During training, some work fine-tunes the encoder to achieve adversarial robustness~\cite{mao2022understanding}, while others demonstrate that zero-shot defense can be significantly improved by fine-tuning the prompts~\cite{li2024one}. However, these methods require additional computational resources and may not be effective against all types of adversarial attacks. On the other hand, test-time adaptation defense methods aim to fit the model during the inference phase without modifying the training process. Methods such as TAPT~\cite{wang2025tapt} and R-TPT~\cite{sheng2025r} tune model parameters during testing to defend against adversarial attacks, thus providing a flexible and label-free solution after deployment. Training-free defense methods take a different approach. Instead of modifying the model itself, they change the adversarial input or its feature embeddings. This technique provides an external defense mechanism that can be applied in real time with minimal computational overhead.
\begin{figure*}[t]
	\centering
    \includegraphics[width=0.9\linewidth]{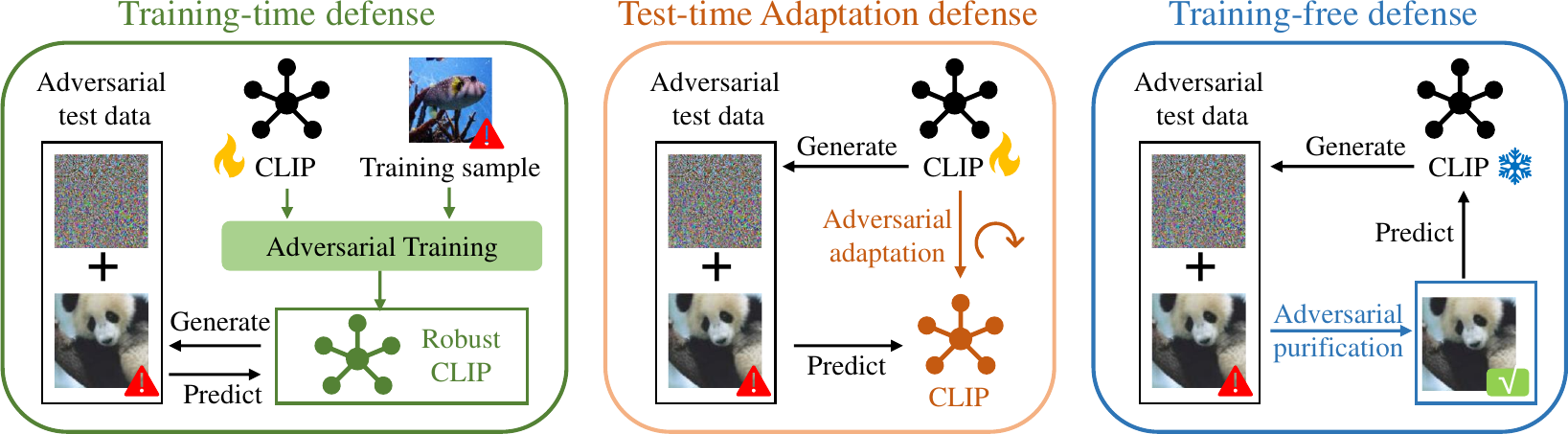}
	\caption{The basic processes of three different defense paradigms. Training-time defense and Test-time Adaptation defense update model parameters during the training and testing phases, respectively, while Training-free defense keeps the parameters unchanged.}
	\label{fig1}
\end{figure*}
In summary, this work makes three contributions. First, it provides a systematic review of adversarial defense strategies for VLM. Second, it synthesizes and categorizes existing research. Third, it identifies some challenges and potential research directions in the field of VLM defense.

The remainder of this survey is organized as follows. Section II introduces the image prediction mechanism of CLIP and the fundamental paradigm of adversarial attacks. Section III provides a comprehensive taxonomy and review of the three main defense paradigms: Training-time Defense, Test-time Adaptation Defense, and Training-free Defense. Section IV presents experimental comparisons of different defense strategies. Finally, Section V concludes the survey and discusses promising future research directions.

\section{Image Prediction and Attack Paradigm of CLIP}

%In this paper, we investigate the zero-shot adversarial generalization capability of VLMs and focus on the basic image classification task. 

Taking the widely-used CLIP as a representative example, let $F_{\theta}(\cdot)$ denote the image encoder of CLIP parameterized by $\theta$, and $G_{\phi}(\cdot)$ represent the text encoder parameterized by $\phi$. Given an input image $x$ and a corresponding category textual description $t$, such as ``This is a photo of $\{\}$'', where $\{\}$ contains the text of class, the image encoder and the text encoder output the corresponding deep feature embeddings $F_{\theta}(x)$ and $G_{\phi}(t)$, respectively. In the context of the image classification task, each category is associated with a specific text description, resulting in a total of $c$ categories. CLIP's prediction is formulated as a retrieval task, where the goal is to identify the text descriptor $G_{\phi}(t_{m})$ that is most similar to the given image embedding $F_{\theta}(x)$ in a multi-modal space, where $t_{m}$ corresponds to the text description of the $m$-th category. The model outputs a $c$-dimensional vector representing the similarity scores, and the category with the highest similarity is selected as the final classification result: \begin{equation}
C\!(x,\!t)\! \!=\! \!\left[\operatorname{sim}\!\left(\!F\!_{\theta}(x), \! G\!_{\phi}\!\left(t_{1}\right)\right),\!  \ldots,\! \operatorname{sim}\!\left(\!F\!_{\theta}(x), \! G\!_{\phi}\!\left(t_{c}\right)\right)\right]
  \label{eq0}
\end{equation}

Based on this, attackers typically seek an imperceptible adversarial perturbation through gradient backpropagation or other methods, and superimpose it onto the original image to trick CLIP into outputting incorrect predictions. Formally, the regular generation process of the perturbation $\delta$ is:
\begin{equation}
 \delta = \max _{\|\delta\|_p \leq \epsilon} {L}(C\left({x}+\delta, t), y \right)
  \label{eq2}
\end{equation} where $\delta$ is within the $\epsilon$-ball (bounded by the $L_p$ norm) centered at the natural input ${x}$, the corresponding label is ${y}$, $L(\cdot)$ represents the supervision loss.

\section{The Taxonomy of Defense Paradigms}

Figure~\ref{fig1} illustrates the basic processes of different types of defense paradigms. Table~\ref{tab1} summarizes the challenges and optimization objects addressed by specific methods. We will describe these methods in detail below.

\subsection{Training-time defense paradigm}

As shown in Figure~\ref{fig1}, the defense paradigm employs a typical adversarial training strategy. Specifically, the adversarial samples are used as training data to update CLIP in a supervised manner to obtain an adversarial robust model. Based on the model parameters that need to be updated, this paradigm can be divided into \textit{Adversarial Fine-tuning (AFT)}~\cite{yu2024text,zhou2024revisiting,wang2025double} and \textit{Adversarial Prompt Tuning (APT)}~\cite{li2024one}. The former typically updates the model's image encoder parameters, making the entire model adversarially robust. The latter enhances the defense capabilities through learnable prompts.

\begin{table*}[t]
\caption{Summary of defense methods for different categories. The table displays the challenges they address and their optimization objects. The code link has been provided.}
\begin{center}
\begin{tabular}{|c|c|c|c|}
\hline
Method& Paradigm Type& Addressed Challenges & Optimized Objects \\
\hline
\href{https://github.com/cvlab-columbia/ZSRobust4FoundationModel}{TeCoA}& Training-time Defense (AFT) & Zero-shot Adversarial Robustness& Image Encoder \\
\hline
\href{https://github.com/serendipity1122/Pre-trained-Model-Guided-Fine-Tuning-for-Zero-Shot-Adversarial-Robustness}{PMG-AFT}& Training-time Defense (AFT) &  Generalization-Robustness Trade-off & Image Encoder \\
 \hline
\href{https://github.com/chs20/RobustVLM}{FARE}& Training-time Defense (AFT) &  Generalization-Robustness Trade-off & Image Encoder \\
\hline
TIMA& Training-time Defense (AFT) &  Generalization-Robustness Trade-off & Image Encoder
 \\
\hline
\href{https://github.com/zhyblue424/TGA-ZSR}{TGA-ZSR}& Training-time Defense (AFT) &  Generalization-Robustness Trade-off & Image Encoder \\
\hline
\href{https://github.com/LixiaoTHU/LAAT}{LAAT}& Training-time Defense (AFT) &  Zero-shot Adversarial Robustness & Image Encoder \\
\hline
\href{https://github.com/ElleZWQ/MMCoA.git}{MMCoA}& Training-time Defense (AFT) &  Robustness under the Multi-modal Attack & Image Encoder, Text Encoder \\
\hline
\href{https://doublevisualdefense.github.io/}{LLaVA}& Training-time Defense (AFT) &  Adversarial Training from Scratch & VLM Parameters \\
\hline
\href{https://github.com/TreeLLi/APT}{APT}& Training-time Defense (APT) & Zero-shot Adversarial Robustness & Text Prompt \\
\hline
\href{https://github.com/jiamingzhang94/Adversarial-Prompt-Tuning}{AdvPT}& Training-time Defense (APT) & Zero-shot Adversarial Robustness & Text Prompt \\
\hline
\href{https://github.com/TomSheng21/R-TPT}{R-TPT}& Test-time Adaptation Defense  & Inference Adversarial Robustness & Text Prompt \\
\hline
\href{https://github.com/xinwong/TAPT}{TAPT}& Test-time Adaptation Defense  & Inference Adversarial Robustness  & Image Prompt, Text Prompt \\
\hline
AOM& Training-free Defense &   Generalization-Robustness Trade-off, Efficiency, Simplicity
& Image Embedding \\
\hline
\href{https://github.com/TMLResearchGroup-CAS/CLIPure}{CLIPure}& Training-free Defense &  Zero-shot Adversarial Robustness, Theoretical Foundation
& Image Embedding  \\
\hline
COLA& Training-free Defense & Adversarial Image-text Misalignment, Theoretical Foundation & Image and Text Embeddings \\
\hline
\href{https://github.com/Sxing2/CLIP-Test-time-Counterattacks}{TTC}& Training-free Defense &   Generalization-Robustness Trade-off, Efficiency, Simplicity
& Test Image  \\
\hline
\end{tabular}
\label{tab1}
\end{center}
\end{table*}
\subsubsection{Adversarial Fine-tuning}

\textbf{TeCoA} firstly achieves defense of large-scale VLMs in zero-shot settings. It enhances the model's resilience by supplementing attack examples into the training data, specifically applying a supervised loss between the adversarial inputs and the text embeddings. 

\textbf{PMG-AFT} focuses on fine-tuning model by leveraging the knowledge already embedded in the pre-trained CLIP. This approach adapts the model to handle adversarial examples more effectively, enhancing its zero-shot natural generalization performance in the presence of adversarial attacks.

\textbf{FARE} demonstrates that supervised fine-tuning struggles to maintain CLIP's performance on natural samples. Therefore, it introduces an unsupervised adversarial fine-tuning scheme that forces perturbed point features to remain close to those of the unperturbed points in the original CLIP.

\textbf{TIMA} introduces Minimum Hyperspherical Energy and Text-distance based Adaptive Margin to increase the inter-class distance between text and image embeddings, respectively. Simultaneously, it utilizes knowledge distillation to maintain embedding consistency before and after fine-tuning. In this way, TIMA achieves a balance between adversarial robustness and generalization ability under large perturbations.

\textbf{TGA-ZSR} proposes enhancing adversarial robustness against attacks by refining text-guided attention. It introduces two components: the Attention Refinement module, which aligns adversarially-induced attention with clean attention, and the Attention-based Model Constraint module, which preserves performance on clean samples while boosting adversarial robustness during training.

\textbf{LAAT} leverages fixed semantic anchors from text encoders for each category, which are used during adversarial training to enhance the robustness of CLIP, particularly for novel categories. However, naively using text encoders initially leads to poor performance due to high cosine similarity in CLIP anchors. To address this, LAAT incorporates an expansion algorithm and alignment cross-entropy loss.

\textbf{MMCoA} explores the impact of both text-based and multi-modal attacks in CLIP. The proposed approach aligns clean and adversarial text embeddings with adversarial and clean visual features through a multi-modal contrastive adversarial training loss, improving the adversarial robustness of both the image and text encoders in CLIP.

\textbf{LLaVA} utilizes an adversarial pre-training paradigm, learning directly from a large amount of data without relying on pre-trained weights. Additionally, it incorporates adversarial visual instruction tuning to further strengthen the defense. The resulting models demonstrate substantial improvements in zero-shot robustness, setting new benchmarks in adversarial defense for VLMs.

\subsubsection{Adversarial Prompt Tuning}

\textbf{APT} focuses on learning a robust text prompt rather than modifying model weights. The method stems from the observation that adversarial defenses are highly sensitive to the choice of text prompt. 

\textbf{AdvPT} leverages learnable text prompts, aligning them with adversarial image embeddings to mitigate vulnerabilities without requiring extensive parameter updates.

\subsection{Test-time adaptation defense paradigm}

Test-time defense~\cite{sheng2025r} generally involves adaptive adjustments to model parameters during the inference phase to enhance adversarial robustness. This technique typically optimizes prompts to align data distributions, ensuring stable performance across adversarial conditions. The key advantage is its flexibility, as it does not require labeled data or extensive retraining, making it highly adaptable for real-world applications.

\textbf{R-TPT} offers a flexible defense that does not require labeled data. It first reformulates the classic marginal entropy objective by eliminating terms that create conflicts under adversarial conditions, leaving only point-wise entropy minimization. R-TPT employs a confidence-driven fusion scheme that integrates inputs from the most trustworthy perturbed variants of the input to bolster its defensive capabilities.

\textbf{TAPT} aims to defense VLMs against visual adversarial attacks. The approach introduces dual-stream prompts (textual and visual) that are optimized per test sample. This optimization simultaneously minimizes multi-view predictive uncertainty and reduces the distribution gap between adversarial and clean examples in an unsupervised setting.

\begin{table*}[ht]
\caption{Compares clean (CLN) vs. robust (ROB) accuracy (\%) on 9 diverse datasets against a PGD adversary. The leading performance on each dataset is indicated in bold.}
\centering
\begin{tabular}{|l|p{0.5cm}|p{0.5cm}|p{0.5cm}|p{0.5cm}|p{0.5cm}|p{0.5cm}|p{0.5cm}|p{0.5cm}|p{0.5cm}|p{0.5cm}|p{0.5cm}|p{0.5cm}|p{0.5cm}|p{0.5cm}|p{0.5cm}|p{0.5cm}|p{0.5cm}|p{0.5cm}|}
\hline
Method & \multicolumn{2}{c|}{Flower102} & \multicolumn{2}{c|}{OxfordPet} & \multicolumn{2}{c|}{DTD} & \multicolumn{2}{c|}{FGVCAircraft} & \multicolumn{2}{c|}{StanfordCars} & \multicolumn{2}{c|}{EuroSAT} & \multicolumn{2}{c|}{ SUN397} & \multicolumn{2}{c|}{Food101} & \multicolumn{2}{c|}{Caltech101} \\ \hline
 & ROB & CLN  & ROB & CLN  & ROB & CLN  & ROB & CLN  & ROB & CLN  & ROB & CLN  & ROB & CLN  & ROB & CLN  & ROB & CLN \\ \hline
CLIP & 1.10 & 65.50 & 1.00 & 87.40 & 3.00 & 40.60 & 0.00 & 20.10 & 0.00 & 52.00 & 0.00 & 42.60 & 1.10 & 58.50 & 0.70 & 83.90 & 14.70 & 85.70 \\ \hline
FARE  & 17.10 & 48.00 & 31.10 & 79.40 & 15.60 & 32.10 & 1.40 & 10.90 & 6.80 & 38.70 & 10.70 & 21.90 & 14.90 & 52.40 & 11.70 & 55.30 & 50.70 & 81.00 \\ \hline
TeCoA & 21.90 & 36.80 & 38.40 & 62.10 & 17.60 & 25.20 & 2.50 & 5.30 & 8.80 & 20.90 & 12.00 & 16.60 & 19.47 & 36.70 & 13.90 & 30.00 & 55.50 & 71.70\\ \hline
AOM  & 36.92 & 62.59 & 51.39 & 85.20 & 23.89 & 38.32 & 10.56 & 15.55 & 25.68 & \textbf{54.78} & 15.57 & 32.02 & 39.46 & 59.56 & 33.94 & 76.13 & 74.65 & 86.52 \\ \hline
PMG  & 23.40 & 37.00 & 41.20 & 65.90 & 15.00 & 21.80 & 2.20 & 5.60 & 11.70 & 25.40 & 12.60 & 18.50 & 22.60 & 38.00 & 18.60 & 36.60 & 61.10 & 75.50 \\ \hline
TTC  & 39.14 & 64.16 & 57.87 & 83.35 & 27.32 & 36.98 & 13.77 & 18.00 & 33.01 & 48.16 & 12.19 & 53.24 & 41.52 & 55.13 & 57.84 & 82.18 & 65.78 & 86.53 \\ \hline
COLA & \textbf{50.40} & \textbf{66.10} & \textbf{77.20} & \textbf{87.90} & \textbf{34.00} & \textbf{41.00} & \textbf{15.60} & \textbf{20.90} & \textbf{35.40} & 54.20 & \textbf{19.20} & \textbf{53.80} & \textbf{45.30} & \textbf{61.90} & \textbf{63.80} & \textbf{84.50} & \textbf{75.30} & \textbf{88.10} \\ \hline
\end{tabular}
\label{tab2}
\end{table*}

\subsection{Training-free defense paradigm}

The general paradigm of this defense strategy involves enhancing model robustness without requiring parameter updates~\cite{zhang2025clipure}. This approach focuses on purifying inputs either in the image domain or feature domain to mitigate the impact of adversarial perturbations. This method is flexible, efficient, and can be applied without altering model architecture.

\textbf{AOM} discovers that Gaussian noise could significantly enhance the model’s zero-shot performance. The noisy adversarial examples are treated as anchors, and the method then finds a path in the embedding space from the adversarial examples to cleaner samples. It improves the generalization abilities and robustness in a training-free manner.

\textbf{CLIPure} addresses the need for a universal defense against adversarial attacks by devising a purification process that generalizes across unseen threat models. Its theoretical foundation characterizes purification risk through the KL divergence between the joint distributions of the purification and attack processes, modeled via bidirectional SDEs.

\textbf{COLA} addresses the adversarial misalignment issue between image and text features by employing an optimal transport based approach. This method is designed to re-establish both global semantic correspondence and local structural coherence between modalities. It achieves this by mapping perturbed image embeddings into the semantic subspace that is defined by the clean text features of the target classes.

\textbf{TTC} proposes a training-free defense to address the issue of ``falsely stable'' images caused by adversarial perturbations. The method develops the potential of CLIP itself to counterattack adversarial images by adding counterattack perturbations during inference, providing a novel defense. 

\section{Experiments}

Comparative experiments are typically conducted on multiple downstream tasks to comprehensively assess the zero-shot adversarial robustness of the defense strategies. Table~\ref{tab2} summarizes the comparative performance of different methods on 9 classification datasets. These datasets encompass a diverse set of domains, such as Caltech101 for general objects, SUN397 for scenes, DTD for textures, EuroSAT for satellite images, and a variety of fine-grained categories including OxfordPet, StanfordCars, Flower102, Food101, and FGVCAircraft. The table shows the results under the same experimental settings. Specifically, CLIP ViT-B/32 is used as the backbone network. The images are preprocessed to a size of 3 $\times$ 224 $\times$ 224 using bicubic interpolation 
followed by channel-wise normalization. The PGD-10~\cite{DBLP:conf/iclr/MadryMSTV18} with $l_{\infty}$ norm perturbation bound of 1 / 255 is used for testing.

Table~\ref{tab2} shows the natural generalization accuracy and adversarial robustness of the original CLIP and 6 representative defense methods. These methods include training-time methods TeCoA, PMG, and FARE; and training-free methods AOM, TTC, and COLA. It can be seen that compared to the adversarial fine-tuning approach, training-free defenses exhibit better performance and demonstrate stronger adversarial robustness potential. Among them, COLA achieves optimal performance on multiple datasets in a more efficient way.

\section{Conclusion and Future Work}

In conclusion, while significant progress has been made in improving the adversarial robustness of VLMs, several challenges still need to be addressed in order to ensure the models are robust enough for real-world applications. There remains a need for more effective solutions that can handle cross-modal adversarial attacks, which target both image and text modalities. Furthermore, more adaptive defense mechanisms are required that can dynamically adjust to different types of adversarial attacks, enhancing the flexibility and robustness of the models. Additionally, future work should focus on improving zero-shot robustness with minimal labeled data, perhaps through semi-supervised or self-supervised learning techniques, to reduce reliance on large annotated datasets. Finally, integrating generative models with adversarial defense strategies may offer new avenues for improving robustness, helping VLMs better withstand a wide range of adversarial perturbations. By addressing these issues, future research will be able to advance the development of more resilient, efficient, and adaptive models, ensuring the security and reliability of VLMs in a variety of real-world scenarios.

\bibliographystyle{plain}
\bibliography{main}

\end{document}